\documentclass[letterpaper, 10 pt, conference]{ieeeconf}  % Comment this line out if you need a4paper

\IEEEoverridecommandlockouts                              % This command is only needed if 
\usepackage{cite}
\usepackage{amsmath, bm}
\usepackage{amssymb,amsfonts}
\usepackage{algorithm}
\usepackage{algpseudocode}
\usepackage{array}
\usepackage{graphicx}
\usepackage{textcomp}
\usepackage{float}
\usepackage{xcolor}
\usepackage{csquotes}
\usepackage{array}
\usepackage{makecell}
\DeclareMathOperator{\E}{\mathbb{E}}
\DeclareMathOperator{\KL}{\mathbb{KL}}

\title{\LARGE \bf
Competency Assessment for Autonomous Agents \\ using Deep Generative Models}

\author{Aastha Acharya$^{1, 2}$, Rebecca Russell$^{2}$ and Nisar R. Ahmed$^{1}$% <-this % stops a space
\thanks{The authors are with $^{1}$Ann and H.J. Smead Department of Aerospace Engineering Sciences at the University of Colorado Boulder, Boulder, Colorado and $^{2}$The Charles Stark Draper Laboratory, Inc., Cambridge, Massachusetts. {\tt\small aastha.acharya@colorado.edu, rrussell@draper.com, nisar.ahmed@colorado.edu}}%
}

\begin{document}

\maketitle
\thispagestyle{empty}
\pagestyle{empty}

%%%%%%%%%%%%%%%%%%%%%%%%%%%%%%%%%%%%%%%%%%%%%%%%%%%%%%%%%%%%%%%%%%%%%%%%%%%%%%%%
\begin{abstract}

For autonomous agents to act as trustworthy partners to human users, they must be able to reliably communicate their competency for the tasks they are asked to perform.
Towards this objective, we develop probabilistic world models based on deep generative modelling that allow for the simulation of agent trajectories and accurate calculation of tasking outcome probabilities.
By combining the strengths of conditional variational autoencoders with recurrent neural networks, the deep generative world model can probabilistically forecast trajectories over long horizons to task completion.
We show how these forecasted trajectories can be used to calculate outcome probability distributions, which enable the precise assessment of agent competency for specific tasks and initial settings. 

\end{abstract}

%%%%%%%%%%%%%%%%%%%%%%%%%%%%%%%%%%%%%%%%%%%%%%%%%%%%%%%%%%%%%%%%%%%%%%%%%%%%%%%%
\section{INTRODUCTION}

With the advancement of machine learning and artificial intelligence, we are approaching an era where autonomous agents can work alongside humans as partners. To get to this stage, however, significant progress still remains to be made in developing safe and trustworthy autonomous agents that are able to work reliably with and under the supervision of humans.
The level of trust a human user has in an autonomous agent influences the user's tasking and usage of the agent under different conditions.
Appropriate trust is particularly critical for autonomous agents based on reinforcement learning (RL), which are typically uninterpretable and can exhibit behavior and performance unaligned with human expectations.

The role of trust in creating effective human-autonomy teaming has been studied extensively \cite{trust_in_automation, buildingtrust}. In particular, it has been shown that the level of trust from the user towards an autonomous agent
is influenced by \emph{assurances} \cite{Aitken2016AssurancesAM, israelsen}, which should be designed to encourage proper usage of these systems. 
One component of assurance is an assessment of the agent's \emph{competency}, its capability in performing a given task, which is impacted by both internal and external factors \cite{cummings}.

\begin{figure}
  \centering
  \includegraphics[scale=0.25]{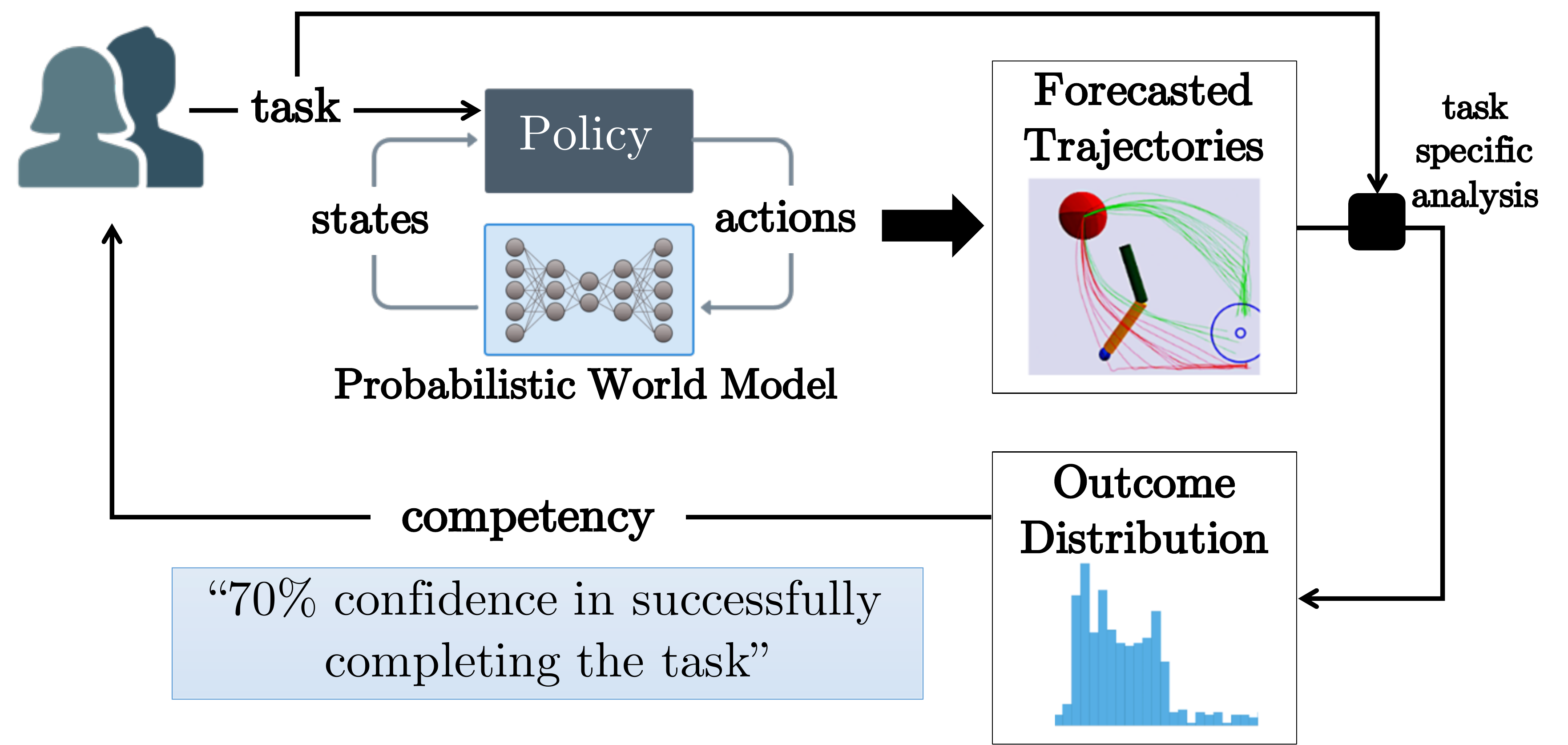}
  \caption{Overview of the proposed method. Given a task by the user, a probabilistic world model is used in conjunction with an autonomous agent's policy to forecast trajectories. These trajectories are abstracted into outcome distributions using task-specific, user-selected criteria. These outcome distributions allow for the formulation and reporting of competency to the user for the provided task.}
  \label{overview}
\end{figure}

In this work, we introduce an approach to competency assessment based on the probabilistic forecasting of trajectories and their outcomes for a given task and initial state condition. Our basic approach, shown in Figure~\ref{overview}, uses a probabilistic world model (PWM) to simulate full trajectories using the autonomous agent's policy in the forecasting loop. These forecasted trajectories are analyzed based on the task to assess the probability distribution of different outcomes and thus the competency of the agent in the scenario. The major technical challenge is in creating a PWM that is able to accurately capture the full joint probability distribution of future states conditioned on the agent's actions over long horizons. To address this challenge, we use deep generative modeling with variational autoencoders (VAEs) to learn the PWM directly from trajectory data.

We contribute (i) a definition of competency and its various components under our problem statement; (ii) a recurrent VAE PWM that is able to capture the high-dimensional, non-Gaussian, and multi-modal probability distributions required for competency assessment via forecasting; (iii) results from experimentation in simulated environments that are deterministic, stochastic, and partially observable. Preliminary material and related works are provided in Section \ref{prelims}. The proposed method with details on our developed model and other comparison models are given in Section \ref{methods}. Results on benchmark RL tasking scenarios for competency assessment are presented in Section \ref{results}. Finally, we conclude with additional discussion points and future work in Section \ref{conclusions}.

\section{BACKGROUND AND RELATED WORK}\label{prelims}

\subsection{Competency and World Models}
Design of competency-aware autonomous agents is a fairly nascent field, especially in the domain of deep RL and autonomous robotics in general. Some previous works have looked at assessing competency bounds for unmanned vehicles \cite{cummings}, estimating the confidence of a robotic system in performing a given task \cite{Kaipa2015TowardET}, and using human feedback to build the agent's competency model over time \cite{Basich_2020}. All of these works are with non-deep learning systems. For deep RL specifically, a number of works have developed approaches for explainability, which is linked to the assessment of competencies. These methods usually perform post-hoc analysis of models and policies \cite{acharya2020explaining, puiutta2020explainable,  heuillet2020explainability} with the intent of explaining the observed behavior. Our focus is instead on developing methods to forecast tasking outcomes for autonomous agents in order to assess their competency. 

To outline the ties between competency, outcomes, and the world model of the autonomous agent, it is important to first clarify the terminology. First, we define the \emph{world model} to be a learned model that predicts the resulting sequence of states conditioned on an initial state and a sequence of actions. Our framework specifically relies on a probabilistic world model (PWM) that not only captures the dynamics of the agent-environment system, but quantifies the uncertainty due to stochasticity or lack of observability. Using this PWM in conjunction with an agent policy lets us simulate full trajectories which we then abstract into a set of \emph{outcomes} based on the tasks that are of interest to the user. These outcomes are quantitative measures of the forecasted trajectories, and hence reflect the agent's competency for a given task. We define the tasking \emph{competency} of an autonomous agent as our belief in its performance and the expected outcomes under given starting conditions. For example, in the case of an autonomous air vehicle that is tasked to fly to a target location, the competency that is reported may contain information such as \enquote{$80\%$ confidence in reaching the target location within 10 minutes.}

The communication of this information by the autonomous agent is an assurance \cite{Aitken2016AssurancesAM, israelsen} that is provided to the user, and we want to provide this information prior to the agent's deployment into the real world. In addition, if the autonomous agent is able to not only communicate whether it can or cannot perform the given task but also indicate what conditions affect its performance, this will help in establishing an appropriate level of trust from the human user. This motivates the need for a robust PWM that is able to efficiently learn the dynamics and perform long-horizon forecasting.

World models have been well-explored in the context of model-based RL. For example, \enquote{World Models}~\cite{ha2018worldmodels} and PlaNet~\cite{hafner2018planet} both showed how a policy can be trained in the learned world model representations of an RL environment. Both of these papers work with high-dimensional image state representations, in contrast to our work that focuses on lower-dimensional vector state spaces. Additionally, these papers are focused on performing efficient reinforcement learning of a policy, which is not the focus of our work. 

In the arena of low-dimensional state representation for model-based RL, previous works have designed models to capture uncertainties that may arise. The uncertainties can be categorized as either epistemic, arising from the model learning process, or aleatoric, irreducible uncertainty. A well-known paper proposed PETS \cite{chua2018deep}, which combines probabilistic world models and ensembles to capture both aleatoric and epistemic uncertainties. Works previous to this had been focused specifically on capturing epistemic uncertainty \cite{nagabandi_mpc}. Even if they are designed to capture uncertainties, these models are not necessarily suited to long-horizon forecasting, which is an important feature of our work. However, as a point of comparison detailed in Section \ref{models_comp}, we take insights from these existing methods to show what a naive approach to competency assessment may look like.

\subsection{Problem Statement}

Since robotics controls applications are of interest, it is assumed the autonomous agent is a complex dynamical system that is interacting with its environment over time. This interaction between the environmental factors and the agent's dynamics produces complex behaviors that are to be captured using the PWM. Therefore, the PWM should capture time history of states and actions, where states may be a combination of the agent's own dynamical states and any useful and available information from the environment. This information is summarized as a trajectory $\mathcal{T}$: 

\vspace*{-\baselineskip}

\begin{equation}
    \mathcal{\bm{T}} = \{\bm{s}_0, \bm{a}_0, \bm{s}_1, \bm{a}_1, \ldots, \bm{a}_{t-1}, \bm{s}_t, \ldots\}
\end{equation}

\vspace*{-4pt}

\noindent where $\bm{s}_t$ and $\bm{a}_t$ represent states and actions at time $t$, respectively. The length of one trajectory constitutes an episode, and the end of an episode is reached when the autonomous agent either successfully achieves its task or runs out of time. Using this trajectory data, outcomes for the corresponding episode can be categorized. This may be a simple categorization as success/failure, or may indicate continuous factors (such as time to success) or any intermediate objective (such as achievement of sequential goals). Thus, the full trajectory plays a vital role in assessing the various outcomes that may be of interest, ultimately informing the competency of the agent. 

Therefore, much of this work relies on forecasting the agent's trajectories and resulting outcomes using the PWM and a given policy. This PWM, parametrized using a deep neural network, initially uses a collection of trajectories with random actions as training data. This ensures a thorough exploration of the environment during the training phase and also allows the PWM to be task-independent. Once trained, the PWM can then be deployed with any policy during the trajectory forecasting process. The goal of the PWM is to learn the following distribution: 

\vspace*{-5pt}

\begin{equation}\label{eq:traj_prob}
    p(\bm{s}_1, \bm{s}_2, \ldots, \bm{s}_{T}|\bm{s}_0; \bm{a}_0, \bm{a}_1, \dots, \bm{a}_{T-1})
\end{equation}

\noindent where $T$ indicates a time horizon of interest to predict to. There are two important points to note about this distribution:

\begin{enumerate}
    \item Only one state observation is received from the true environment (denoted $\bm{s}_0$), and the rest of the states have to be forecasted without receiving any other environmental observation;
    \item The actions may result from any policy.
\end{enumerate}
Thus, the PWM allows us to assess agent competency using only the initial conditions and one or more policies.

\subsection{Deep Generative Models} \label{dgm}

Deep generative models approximate the joint probability distribution underlying a training distribution. By learning a mapping of the training samples to an  easy-to-sample-from distribution, these models can be used to generate new samples whose distribution approximately matches the training distribution. There are many variations of deep generative models \cite{dgm_survey}, such as generative adversairal networks (GANs) \cite{gans}, flow-based models \cite{normalizing-flows-pmlr-v37-rezende15}, and variational autoencoders (VAEs) \cite{Kingma2014}. GANs use two neural networks, one as a generator and other as a discriminator, in a minimax game objective setting. While GANs are very popular in specific application areas such as computer vision \cite{Karras_2020_CVPR, Radford2016UnsupervisedRL}, they often suffer from mode collapse of the output distribution and are ill-suited to applications that require careful uncertainty quantification. Flow-based models are grounded on the normalizing flows method \cite{Kobyzev2021NormalizingFA}, where sequential invertible transformations are used to directly learn the underlying probability distribution. This method requires computations of the Jacobian determinant, which can lead to excessive computation time, scalability issues, and restrictions on the model architecture.

VAEs emerge from classic variational inference techniques that introduce latent variables into the model to capture any unobserved or salient features of the data. The resulting latent variable space is highly flexible since it can be either continuous and discrete, and because we can impose different forms of prior on the distribution \cite{oord2018neural}. This makes VAEs efficient in capturing diverse set of distributions. Additionally, they can be easily modified to introduce a recurrent structure, which makes them appropriate for dynamics learning over longer horizons. Hence, we choose VAE as the PWM in this work. More specifically, we work with conditional VAE where the objective is to maximize the likelihood of $p\left(y|x\right)$. This likelihood can be expanded using a form that marginalizes out the latent variable $z$ such that $p\left(y|x\right) = \int p\left(y|z,x\right) p\left(z|x\right) \, dz$ so that the training objective contains both the latent variables and the input-output pairs. We expand on this likelihood objective to contain recurrent structure to capture the full time-varying dynamics of the joint probability distribution.

Extensions of VAEs in temporal frameworks have been used in application areas such as text generation \cite{bowman2016generating}, music generation \cite{bayer2015learning}, and image and sketch generation \cite{drawnetworl_gregor, ha2017neural}. These applications tend to have single-dimensional input and do not transfer well to multi-dimensional systems. There are also papers that have analyzed temporal VAEs for dynamical model learning. One work has used z-forcing \cite{zforcing} techniques to force the latent space to contain information from the future states in order to predict long-term dynamics\cite{rosemaryke_longtermfuture}. This model is designed to predict the action sequence alongside the states, which puts a task- and policy-specific restriction that is undesirable for competency assessment. Other methods such as temporal-difference VAEs \cite{gregor2019temporal} allow for jumpy predictions that skip prediction of the middle states. From a competency assessment perspective, all of the states in a trajectory are valuable for extracting intermediate outcome probabilities. Moreover, in comparison to these existing works, the strength of our method lies in its simplicity and ease of implementation while still producing a robust probabilistic prediction model.

\section{METHODS}\label{methods}

\subsection{Model Development}\label{model_dev}

\begin{figure}
  \centering
  \includegraphics[scale=0.4]{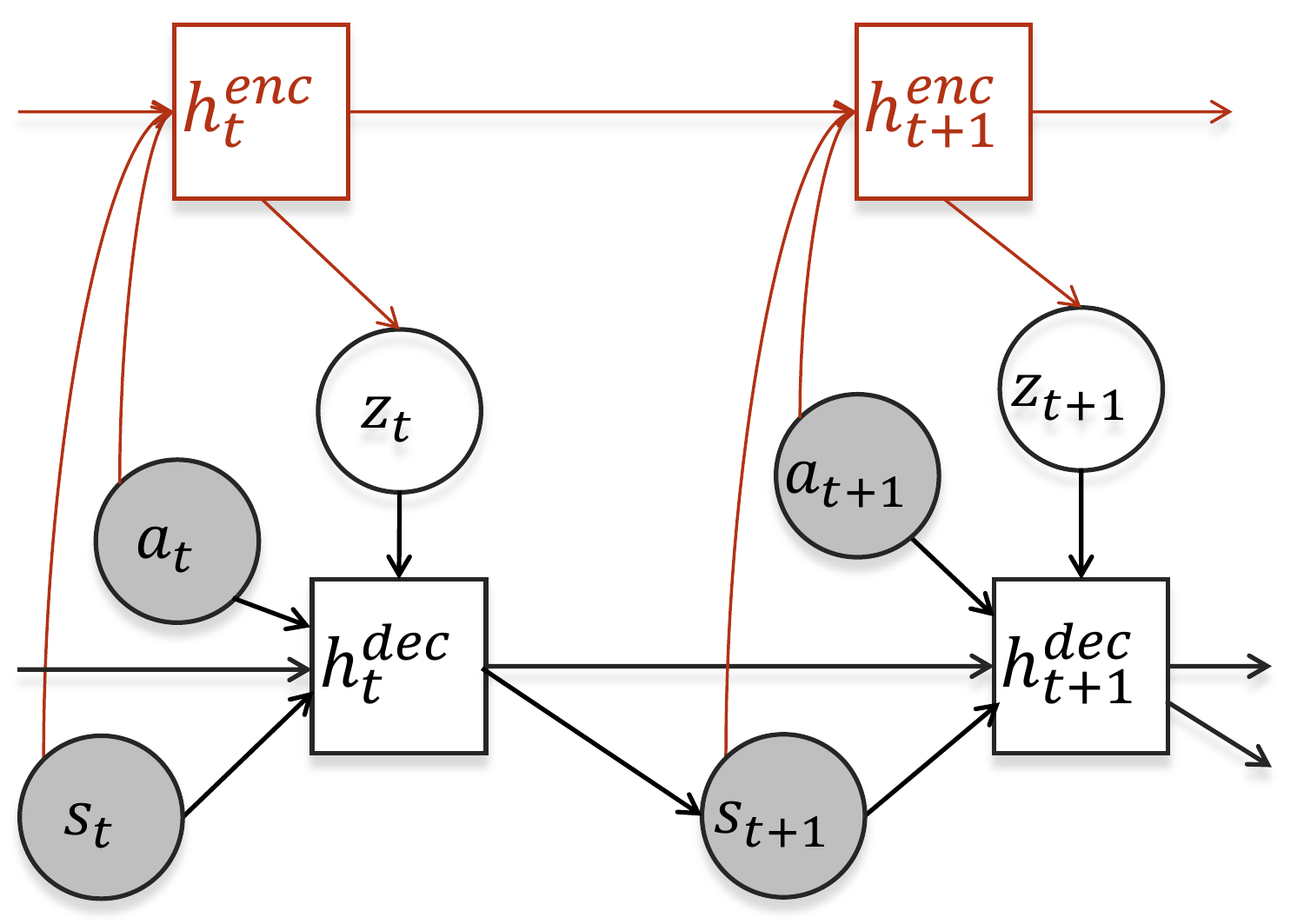}
  \caption{Graphical model showing stochastic (circles) and deterministic (squares) variables and their connections. Variables that are observed during training are shaded. Connections and variables in red show how the latent variable is informed during the training process, and are otherwise discarded during sample generation process.}
  \label{pgm}
\end{figure}

% \vspace*{-\baselineskip} 

The objective of the PWM training is to maximize the likelihood of the distribution provided in Equation \ref{eq:traj_prob}, where the expanded autoregressive form contains the product of the following individual distributions: 

\vspace*{-\baselineskip} 

\begin{equation}
\begin{split}
p(\bm{s}_1, \bm{s}_2, &\ldots, \bm{s}_{T}|\bm{s}_0; \bm{a}_0, \bm{a}_1, \dots, \bm{a}_{T-1}) = \\
    &p\left(\bm{s}_{1}|\bm{s}_0; \bm{a}_0\right) \times \\
    &p\left(\bm{s}_{2}|\bm{s}_{0}, \bm{s}_{1}; \bm{a}_{0}, \bm{a}_{1}\right) \times \\
    & \qquad \vdots \\
    \times \quad &p\left(\bm{s}_{T}|\bm{s}_{0}, \ldots,\bm{s}_{T-1}; \bm{a}_{0}, \ldots, \bm{a}_{T-1}\right)
\end{split}
\end{equation}

This form allows latent variables to be introduced into each of the terms in the product as follows: 

\vspace*{-\baselineskip}

\begin{equation}
\begin{split}
    &p\left(\bm{s}_{1}|\bm{s}_0; \bm{a}_0\right) = \\
    & \qquad  \int p\left(\bm{s}_{1}|\bm{s}_0; \bm{a}_0; \bm{z}_0\right) \times p\left(\bm{z}_0|\bm{s}_0; \bm{a}_0\right) \, d\bm{z}_0
    \\
    &p\left(\bm{s}_{2}|\bm{s}_{0}, \bm{s}_{1}; \bm{a}_{0}, \bm{a}_{1}\right) = \\
    & \qquad  \int p\left(\bm{s}_{2}|\bm{s}_{0}, \bm{s}_{1}; \bm{a}_{0}, \bm{a}_{1}; \bm{z}_{1}\right) \\
    & \qquad  \times p\left(\bm{z}_{1}|\bm{s}_{0}, \bm{s}_{1}; \bm{a}_{0}, \bm{a}_{1}\right) \, d\bm{z}_{1}
    \\
    & \qquad \vdots 
    \\
    &p\left(\bm{s}_{T}|\bm{s}_{0}, \ldots,\bm{s}_{T-1}; \bm{a}_{0}, \ldots, \bm{a}_{T-1}\right) = \\
    & \qquad  \int p\left(\bm{s}_{T}|\bm{s}_{0}, \ldots,\bm{s}_{T-1}; \bm{a}_{0}, \ldots, \bm{a}_{T-1}; \bm{z}_{T-1}\right) \\
    & \qquad  \times p\left(\bm{z}_{T-1}|\bm{s}_{0}, \ldots,\bm{s}_{T-1}; \bm{a}_{0}, \ldots, \bm{a}_{T-1}\right) \, d\bm{z}_{T-1}
    \label{eq:latent_var_expanded}
\end{split}
\end{equation}

The graphical model of this representation is seen in Figure \ref{pgm}, where $\bm{h}$ represents recurrency. The two distinct recurrencies for the encoder ($\bm{h}^{enc}$) and the decoder ($\bm{h}^{dec}$) have the following forms:

\vspace*{-\baselineskip}

\begin{equation}
    \begin{split}
        \bm{h}^{enc}_{t} &= f\left(\bm{s}_{t}, \bm{a}_{t}, \bm{h}^{enc}_{t-1}\right) \\
        \bm{h}^{dec}_{t} &= f\left(\bm{s}_{t}, \bm{a}_{t}, \bm{z}_{t}, \bm{h}^{dec}_{t-1}\right)
    \end{split}
\end{equation}

Using this formulation, the latent variable achieves the following desired distribution: 

% \vspace*{-\baselineskip}

\begin{equation}
    p\left(\bm{z}_{t}|\bm{h}^{enc}_{t}\right) = p\left(\bm{z}_{t}|\bm{s}_{0}, \ldots, \bm{s}_{t}; \bm{a}_{0}, \ldots, \bm{a}_{t}\right)
\end{equation}

Then, the prediction at the next time step can also be written in a similar manner: 

\vspace*{-\baselineskip}

\begin{equation}
    p\left(\bm{s}_{t+1} | \bm{h}^{dec}_{t}\right) = p\left(\bm{s}_{t+1}|\bm{s}_{0}, \ldots, \bm{s}_{t}; \bm{a}_{0}, \ldots, \bm{a}_{t}; \bm{z}_{t}\right)
\end{equation}

Returning to the original maximum likelihood objective in Equation \ref{eq:traj_prob} and taking the logarithm, we get the following: 

\vspace*{-\baselineskip}

\begin{equation}
\begin{split}
    \log p\left(\bm{s}_{1}, \ldots, \bm{s}_{T}|\bm{s}_0; \bm{a}_0, \dots, \bm{a}_{T-1} \right) = \\
    \sum_{t = 0}^{T-1} \log p\left(\bm{s}_{t+1}|\bm{s}_{0:t} ; \bm{a}_{0:t}\right)
\end{split}
\end{equation}

% \vspace*{-\baselineskip}

\begin{figure}
  \centering
  \includegraphics[scale=0.4]{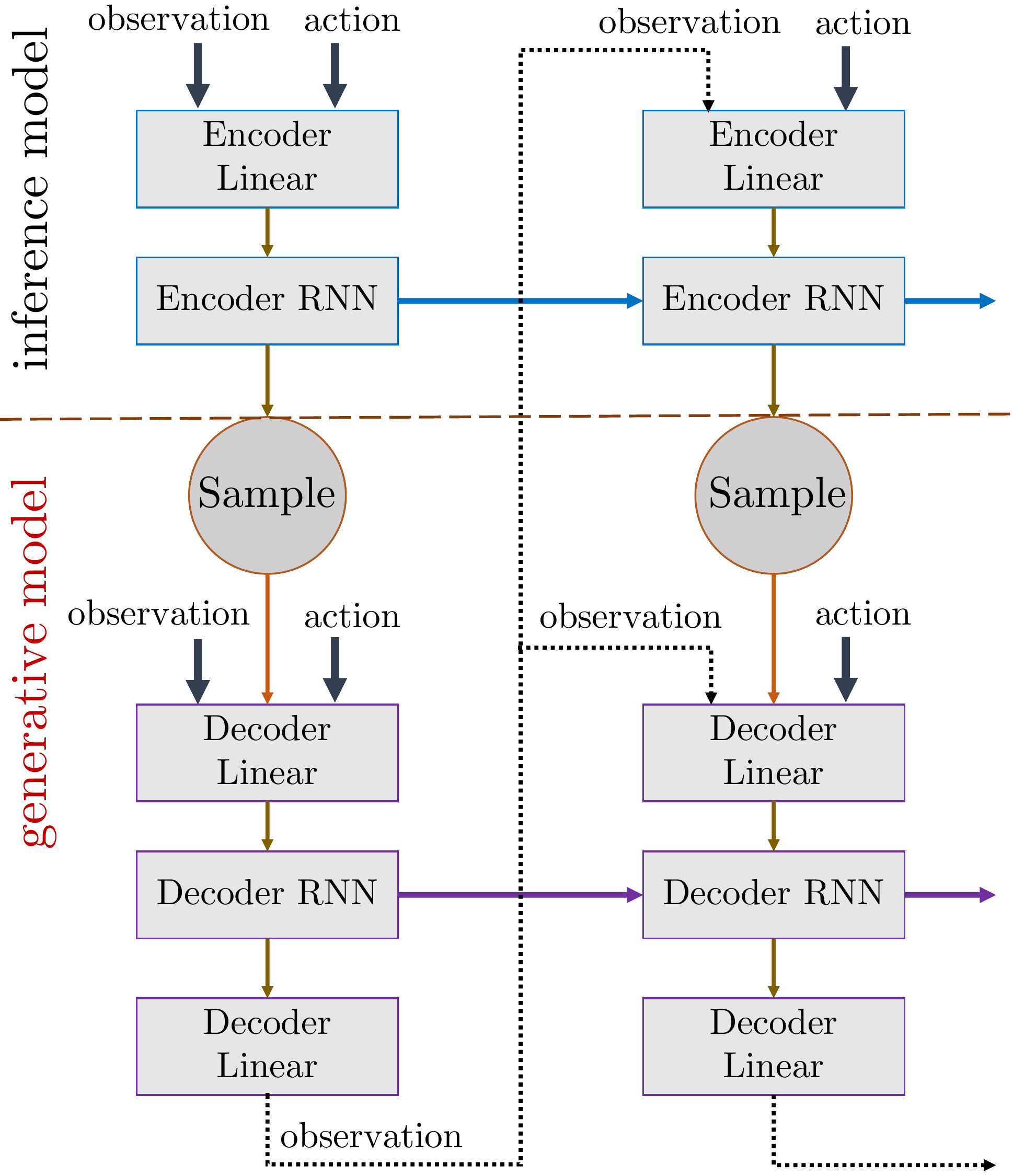}
  \caption{Model architecture showing connections between encoder, sampling, and decoder layers. The encoder layers make up the inference model and are only used during training. Sampling and decoder layers make up the generative model and are used to generate new trajectory samples.}
  \label{frmwork}
\end{figure}

Hence, each individual term can be considered its own maximization objective criteria. Using the latent variable based form of each term as shown in Equation \ref{eq:latent_var_expanded}, Jensen's inequality and the Evidence Lower Bound (ELBO) formulation as in the original VAE loss \cite{Kingma2014} can be used to achieve the following form of the objective criteria for each individual term of the summation above: 

\vspace*{-\baselineskip}

\begin{equation}
\begin{split}
    &p\left(\bm{s}_{t+1}|\bm{s}_{0:t} ; \bm{a}_{0:t}\right) \geq \\
    & \quad \E_{q_{\phi}\left(\bm{z}_{t}|\bm{s}_{0:t}; \bm{a}_{0:t} \right)}\left[\log p_{\theta}\left(\bm{s}_{t+1}|\bm{s}_{0:t} ; \bm{a}_{0:t} ; \bm{z}_{t} \right)\right] \\
    & \quad - \KL[ q_{\phi}\left(\bm{z}_{t}|\bm{s}_{0:t} ; \bm{a}_{0:t}\right) || p_{\theta}\left(\bm{z}_{t}\right)]
\end{split}
\end{equation}

\noindent where $\phi$ represents the encoder parameter, $\theta$ represents the decoder parameter, and $q_{\phi}$ represents an approximate distribution over the latent variable. The terms shown on the right hand side are known as the reconstruction loss and the KL divergence loss, respectively. The reconstruction loss ensures the output prediction matches the training data, while the KL divergence loss ensures the output latent variable distribution matches our specified prior.

To improve the training, we add one more parameter to the loss function to control the interplay of the two loss terms. It is possible for VAEs to undergo mode collapse, which can be attributed to one term of the loss dominating over the other \cite{lucas2019dont, dai2019usual}. To address this, KL annealing and KL cycling have shown to be effective in ensuring that the \enquote{KL vanishing} issue does not arise \cite{bowman2016generating, cyclickl}. In these methods, the weight on the KL term is gradually increased from $0$ to $1$ throughout the training process. Additionally, there are demonstrable benefits to outweighing the KL term compared to the reconstruction term to ensure proper disentanglement and separation within the latent space \cite{Higgins2017betaVAELB}. With these insights, we use $\beta$ as the weight variable for the KL divergence term, and adjust the value of this parameter from $0$ to a value greater than $1$ during the training process. The final loss function takes on the following form: 

\vspace*{-\baselineskip}

\begin{equation}
\begin{split}
    &\mathcal{L}\left(\bm{s}_{1:T}|\bm{s}_0, \bm{a}_{0:T-1}\right) = \\
    & \quad \sum_{t = 0}^{T-1} - \E_{q_{\phi}\left(\bm{z}_{t}|\bm{s}_{0:t}; \bm{a}_{0:t}\right)}\left[\log p_{\theta}\left(\bm{s}_{t+1}|\bm{s}_{0:t} ; \bm{a}_{0:t}; \bm{z}_{t}\right)\right] \\
    & \quad + \beta \times \KL\left[ q_{\phi}\left(\bm{z}_{t}|\bm{s}_{0:t} ; \bm{a}_{0:t}\right) || p_{\theta}\left(\bm{z}_{t}\right)\right]
\end{split}
\end{equation}

% \vspace*{-5}

The final structure of recurrent VAE model is shown in Figure \ref{frmwork}. The encoder and decoder portions are separated into \enquote{inference model} and \enquote{generative model}, respectively. During the training process, both parts of the model are used, but we are able to discard the inference model during the sample generation process used in forecasting. 

\subsection{Forecasting Methods}

Forecasting trajectories through simulation with a learned PWM occurs with the assumption that we do not have access to real environment beyond the first received observation $\bm{s}_0$. Therefore, the PWM is used as a stand-in for the environment to propagate the states forward. The forecasting process relies on having access to a policy that provides the next action given the current state. For our purposes, we use a filtering and reward-weighted refinement planner \cite{nagabandi_mpc} in a model predictive control framework to generate the actions. This planning method also requires a planning model. Since planning with probabilistic state predictions, as would arise if using the PWM as the world model, is out of scope for this work, we instead use a deterministic RNN model for this purpose. Additionally, given a initial starting state, we generate $N$ trajectory forecasts from the model to create a probabilistic distribution over the trajectories. With this information, the method of forecasting for competency assessment via a PWM is described in Algorithm \ref{alg:forecasting}. 

\begin{algorithm} 
{\small{
	\caption{Trajectory Forecasting and Outcome Assessment } 
	\label{alg:forecasting}
	\begin{algorithmic}[1]
	    \State Input real observation $s_0$ from environment
	    \For {sample $n = 1$ to total samples $N$}
    	    \For {time $t = 0$ to task horizon $T-1$}
    	        \State Use planner to generate $M$ sequences of actions up to 
    	        \State planning horizon $h$ $\bm{a}_{t:t+h}^{1:M}$
    	        \For {sampled actions $\bm{a}_{t:t+h}^{1:M}$}
    	            \State Use planning model to propagate $\Bar{\bm{s}}_{t:t+h}^{1:M}$ 
    	            \State Evaluate using reward function $r(\Bar{\bm{s}}_{t:t+h}^{1:M}, \bm{a}_{t:t+h}^{1:M})$
    	            \State Return first action $\bm{a}^*_t$ from best action sequence
    	       \EndFor
    	       \State Forecast using PWM $\hat{\bm{s}}_{t+1} \sim p(\bm{s}_{t+1} | \bm{s}_0, \hat{\bm{s}}_{1:t}, \bm{a}^*_{0:t})$
    	   \EndFor
	   \State Return forecast $\mathcal{\bm{T}}_{\textit{forecast}} = \{\bm{s}_0, \bm{a}^*_0, \hat{\bm{s}}_{1}, ..., \bm{a}^*_{T-1}, \hat{\bm{s}}_{T}\}$
	   \EndFor
	   \State Abstract collected forecast $\mathcal{\bm{T}}_{\textit{forecast}} = \{\bm{s}_0,\bm{a}^{*1:N}_{0:T-1}, \hat{\bm{s}}_{1:T}^{1:N}\}$ into outcomes using any criteria of interest
	\end{algorithmic} 
	}}
\end{algorithm}

% \vspace*{-\baselineskip}

\subsection{Comparison Models}\label{models_comp}

To benchmark the developed recurrent VAE PWM for competency assessment, we train two additional models that are commonly found in the literature. These models are used for comparison and to demonstrate how they may be used for the purpose of competency assessment. They are: 

\begin{itemize}
    \item \textbf{Deterministic RNN}: A deterministic recurrent neural network that is trained with the same depth and comparable number of parameters to the recurrent VAE model. Using this model results in a binary outcome assessment that reflects the most-likely expected outcome. 
    \item \textbf{Probabilistic MLP} \cite{kendall_gal}: A probabilistic multi-layer perception that outputs mean and variance of a Gaussian distribution at each time step, where the next state is then sampled from this distribution. In the PETS \cite{chua2018deep} framework, this model is used to capture aleatoric uncertainty. However, this model is not designed for long-horizon forecasting. Comparison against this model, which does not have the memory-based training, demonstrates how memory is necessary for forecasting, especially when encountering uncertainties which make the environment non-Markovian. 
    
\end{itemize}

\section{EXPERIMENTS} \label{results}

We perform competency assessment experiments with different variations (deterministic, stochastic, and partially observable) each on two different RL environments: CartPole and Pusher.

The output outcome probabilities of success are analyzed using the Brier score, a proper scoring rule that measures the accuracy of predicted probabilities. It is the mean squared of error of the forecast and is computed as $1/N \sum_{n=1}^N (f_n - o_n)^2$, where $f_n$ is forecasted probability, $o_n$ is the real environmental outcome, and $n=1:N$ denotes the number of forecasting instances. 

\subsection{CartPole Environment}

The CartPole environment consists of a 4-dimensional state space, made up of cart position and velocity $(x, \dot{x})$ and pole angle and angular velocity $(\theta, \dot{\theta})$. The input action is force $F$ applied to the cart that moves it in either direction in 1D space. The success criterion for competency is based on the pole's ability to maintain its upright position for the duration of an episode. We further add complexity into this environment by injecting noise via the actions to create a stochastic environment. The injected noise is Gaussian with standard deviation equal to 0--20\% of the range of allowable action. 

\begin{table}[hbt!]
\caption{Brier score for CartPole environment (lower is better)} 
\begin{center}
\renewcommand{\arraystretch}{1.2}
\begin{tabular}{  | m{1.75cm} || m{1.5cm} | m{1.5cm} | m{1.5cm} | }
 \hline
    & \makecell{Recurrent \\ VAE} & \makecell{Probabilistic \\ MLP} & \makecell{Deterministic \\ RNN}\\
    \hline \hline
    \makecell{CartPole \\ Average} & $\bm{0.145}$  & $0.218$ & $0.312$ \\
    \hline \hline
    \makecell{CartPole \\ Deterministic} & $\bm{0.118}$  & $0.123$ & $0.139$ \\
    \hline
    \makecell{CartPole \\ Stochastic} & $\bm{0.175}$  & $0.313$ & $0.485$ \\
\hline
\end{tabular}
\end{center}
\label{cartpole_brier_scores}
\end{table}

% \vspace*{-\baselineskip}

The results for the three trained models over $1000$ forecasting instances are presented in Table \ref{cartpole_brier_scores}, where lower Brier score is better. The recurrent VAE model outperforms the other two models and is clearly better suited for competency assessment in the CartPole environment. This is also corroborated by the break down of the results based on deterministic and stochastic environment, both of which show recurrent VAE as the best model choice. 

\subsection{Pusher Environment Results}

The Pusher environment is a modified version of OpenAI gym Reacher environment \cite{brockman2016openai} that introduces added complexity and interesting behaviors. While keeping the same double pendulum setup (called the arm), a ball is added that has to be pushed to random target locations. This environment consists of a 12-dimensional state space that is made up of the end of the arm's position in $(x, y)$, the cosine and sine of the upper arm angle, angular rate of the upper arm, the cosine and sine of angle of the lower arm relative to the upper arm, angular rate of the lower arm, and the ball positions and velocities in $(x, y, \dot{x}, \dot{y})$. The action $\bm{T}$ applied is two dimensional and represents the torque applied to the upper arm joint and the lower arm joint. The success criteria here are based on the arm's ability to get the ball to the desired target location while maintaining a low velocity. Since this is a much richer environment than CartPole, we primarily use this environment to showcase several key results. To demonstrate the full capacity of the PWM in performing competency assessment, we first add complexity into this model in two ways, both of which introduce aleatoric uncertainties into this otherwise deterministic world: 

\begin{itemize}
    \item \textbf{\emph{Stochastic Pusher: }} Similar to the CartPole environment, we make the Pusher environment stochastic by injecting noise via the actions. The injected noise here is also Gaussian noise with standard deviation equal to 0--20\% of the range of allowable actions.
    \item \textbf{\emph{Partially Observable Pusher: }} To introduce an unobserved variable, the strength of the arm is varied from 50--100\% during the course of trajectory collection. This information is hidden from the PWM, thus introducing an unobserved variable that is not in the state representation but which directly impacts the dynamics.
\end{itemize}

\begin{table}[hbt!]
\caption{Brier scores for Pusher environment (lower is better)}
\begin{center}
\renewcommand{\arraystretch}{1.2}
\begin{tabular}{  | m{1.75cm} || m{1.5cm} | m{1.5cm} | m{1.5cm} | }
 \hline
    & \makecell{Recurrent \\ VAE} & \makecell{Probabilistic \\ MLP} & \makecell{Deterministic \\ RNN}\\
    \hline \hline
    \makecell{Pusher \\ Average} & $\bm{0.088}$  & $0.199$ & $0.162$ \\
    \hline \hline
    \makecell{Pusher \\ Deterministic} & $\bm{0.084}$  & $0.246$ & $0.139$ \\
    \hline
    \makecell{Pusher \\ Stochastic} & $\bm{0.119}$  & $0.140$ & $0.206$ \\
    \hline
    \makecell{Pusher \\ Partially \\ Observable} & $\bm{0.060}$  & $0.211$ & $0.141$ \\
\hline
\end{tabular}
\end{center}
\label{brier_scores}
\end{table}

% \vspace*{-\baselineskip}

The Brier scores are presented in Table \ref{brier_scores}. The probabilities that are being analyzed are whether the arm is able to get the ball to the target location during a given episode. The results are provided for each of the three trained world models over $1000$ starting conditions, and we also provide a break down of the results based on the three environmental conditions. In all cases, recurrent VAE model has a clear advantage over the two methods in producing accurate probabilities for the purpose of competency assessment. In the deterministic environment, RNN is the next best option, and the higher Brier score value of the probabilistic MLP may shed light to the fact that it is not necessarily designed for forecasting. 

For the stochastic case, the injected noise is Gaussian as is the distribution that is modeled by the probabilistic MLP, which may contribute to probabilistic MLP showing better results than the deterministic RNN, the latter of which understandably performs the worst. Finally, in the partially observable case, both the recurrent VAE and deterministic RNN outperform the probabilistic MLP, which points to the importance of carrying memory in such situations. We emphasize the ability of the recurrent VAE to maintain its low Brier score across all three cases, which further indicates that it is flexible in capturing different forms of uncertainty that affects the agent's competency.

\begin{figure}[ht]
  \centering
  \includegraphics[scale=0.525]{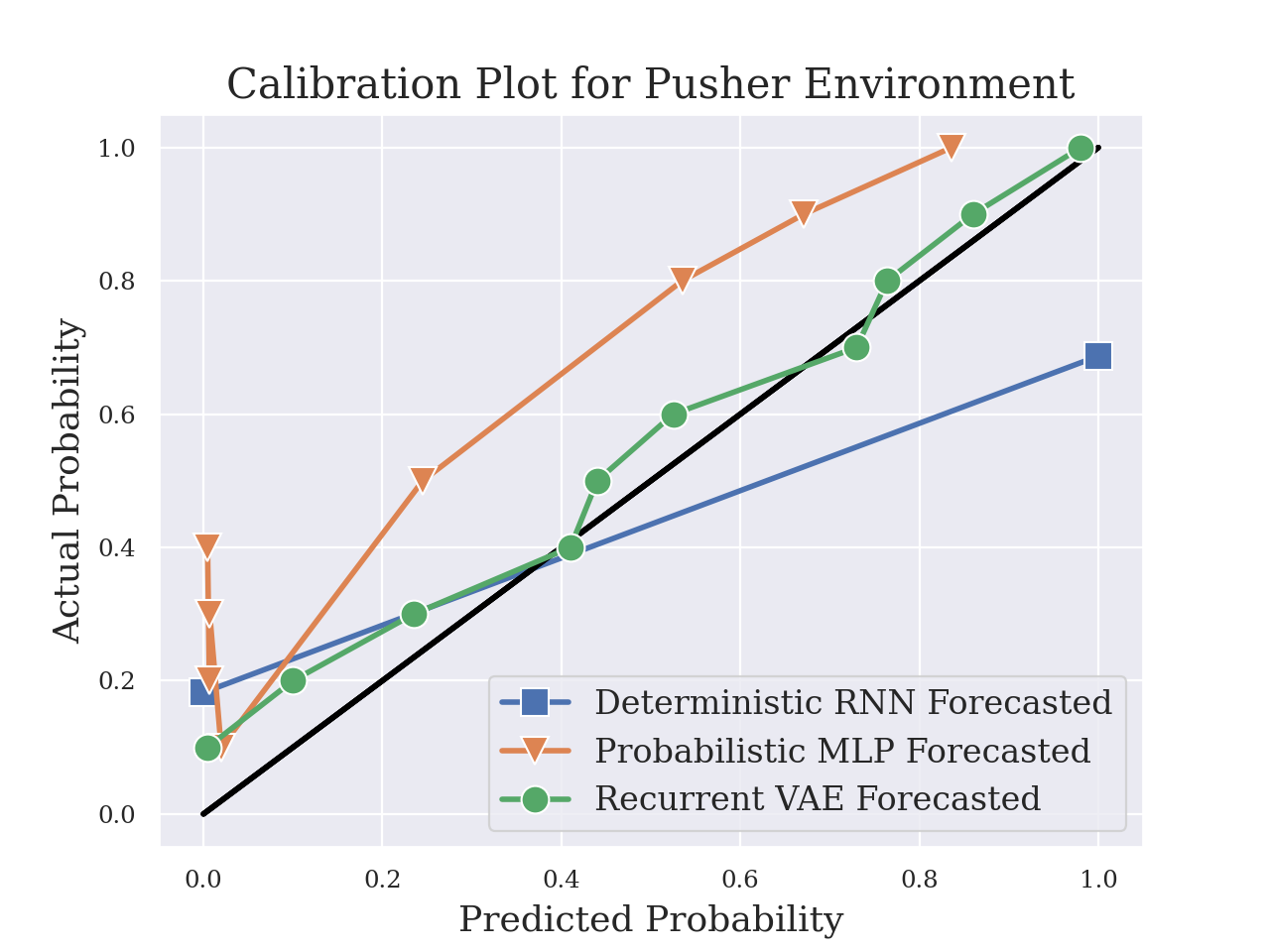}
  \vspace*{-\baselineskip}
  \caption{Calibration curve for the pusher environment (along the black diagonal line is best calibrated).}
  \label{pusher_calib}
\end{figure}

To further expand on this, while Brier scores provide insight on the accuracy of the forecast, probability calibration curve can be used to assess it reliability. Also called the reliability diagram \cite{calibrated_prob}, the calibration curves plot the frequency of what was observed in the environment against the frequency of the predicted probabilities. The results for the Pusher environment are provided in Figure \ref{pusher_calib}, and it is desired for the model to follow the black diagonal line, indicating that it is well calibrated. From the figure, it is clear that the VAE most closely follows the desired calibration level. We expect the deterministic RNN to be only capable of producing success ($1$) or failure ($0$) outcomes, and that is exactly what is observed. Furthermore, the probabilistic MLP is not calibrated with the true probabilities and appears skewed toward producing probabilities on either end, as shown by the lack of points towards the center in the curve for this model. Although not perfect, the recurrent VAE is most closely calibrated in comparison to the other models. 

\begin{figure*}
  \centering
  \includegraphics[scale=0.5]{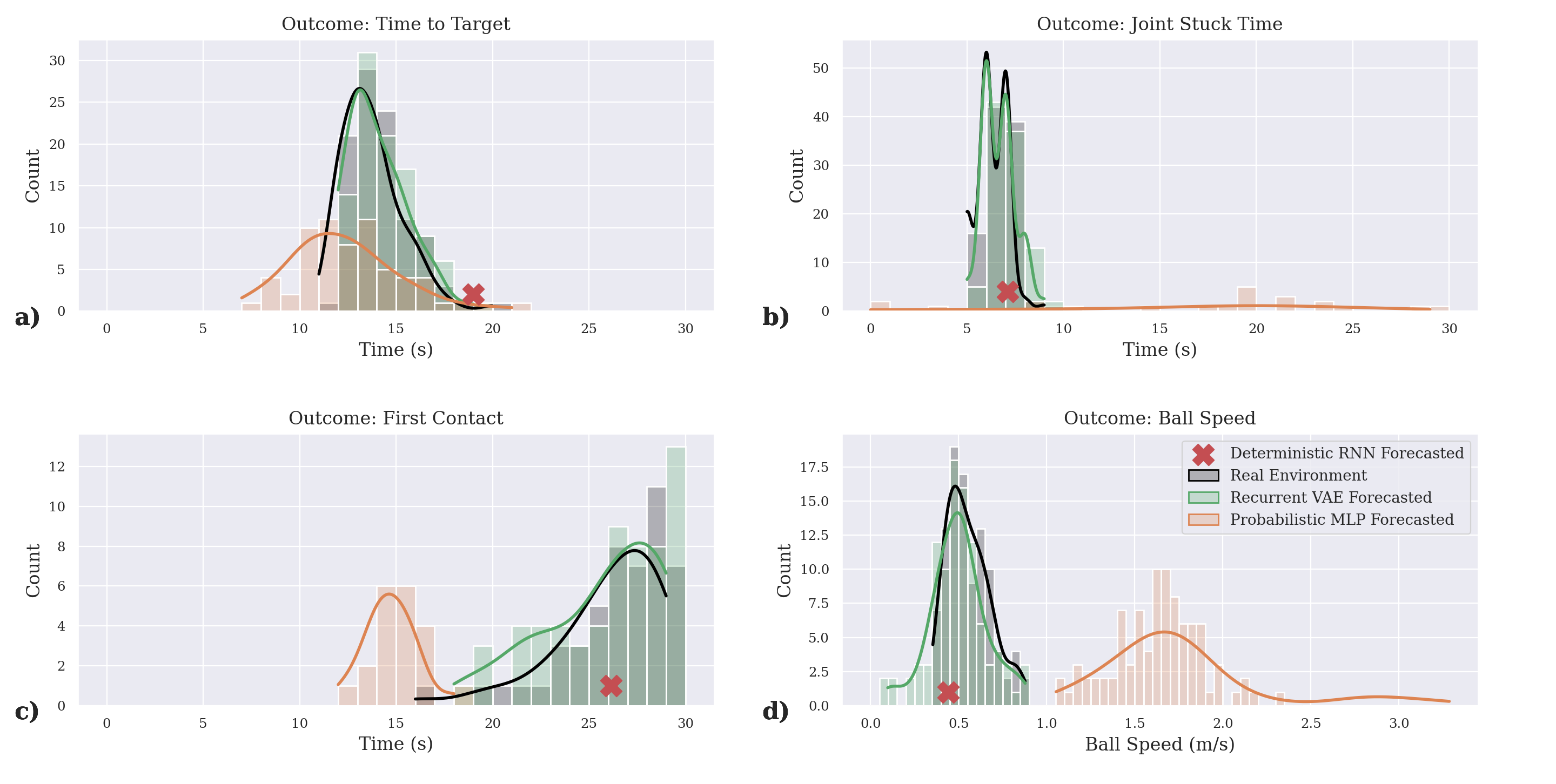}
  \caption{Example of four outcomes that may of interest to the user and the resulting histograms for the Pusher environment. a) Interested outcome: time for the ball to reach target location. b) Interested outcome: total amount of time during an episode where the joint may become stuck. c) Interested outcome: time when the first contact between the arm and the ball occurs. d) Interested outcome: maximum speed reached by the ball during an episode.}
  \label{outcome_hist}
\end{figure*}

\textbf{Multi-Variate Outcome Analysis} One of the benefits of having access to the forecasted trajectories in the process of competency assessment is that we can further analyze intermediate outcomes or other dimensions of interest of the episode. We do this by abstracting the forecasted trajectories into outcome distributions based on criteria that may of interest to the user. For example, in the Pusher case, along with analyzing the overall success/failure, a user may be interested in the amount of time to success or the maximum speed the ball reaches during an episode. Examples of outcome distributions are presented in the Figure \ref{outcome_hist}, where we show histograms that result from the original, non-deterministic environment and as predicted by the models over $100$ runs. 

The four outcomes that we choose to analyze are the time needed to get the ball to target (Figure \ref{outcome_hist}a), amount of time the joint of the arm may be stuck (Figure \ref{outcome_hist}b), time of first contact between the arm and the ball (Figure \ref{outcome_hist}c), and the maximum ball speed reached during a trajectory (Figure \ref{outcome_hist}d). The outcomes presented are for one initial setting of the environment and the outcomes are abstracted from $100$ forecasted trajectories. For a deterministic RNN, the output outcome is always the same for all $100$ runs, hence this value is presented as a single marker point in the plot. Here, we are truly able to observe the efficiency of the recurrent VAE in capturing varying forms of outcome distributions. The output from the deterministic RNN typically tends to be aligned with the average of the real and recurrent VAE forecasting. As for the the probabilistic MLP, we see that the distribution that it produces can be misaligned with the true distribution, which may explain its higher Brier scores than other two models. All of these results combined demonstrate the ability of VAEs to capture both the overall success or failure as well as the intermediate states, which may be of interest to the user for evaluating and communicating different aspects of task competency.

The outcome distributions presented here can also be used to form competency statements for a given task. For example, if asked \enquote{will the ball get to target location within $15$ seconds?} with the initial setting as shown in Figure \ref{outcome_hist}a, the competency statement based on the outcome probability is \enquote{$83\%$ confidence in ball reaching target within $15$ seconds}. Similarly, the user may ask \enquote{will the ball exceed a speed of $0.8$ m/s?} for the scenario in Figure \ref{outcome_hist}d. We can generate a competency statement of the form \enquote{$4\%$ chance of the ball reaching a speed greater than $0.8$ m/s}. These statements will vary based on the initial setting along with the task that is used to the set the policy, and hence can be utilized by the user for a variety of purposes. This shows how these statements can provide assurances that will establish an appropriate level of trust from the user towards the autonomous agents. 

\section{CONCLUSIONS AND FUTURE WORK}\label{conclusions}
As a step towards creating autonomous agents that can act as effective collaborators with humans, we are interested in building mechanisms for automated competency assessment. We envision being able to probablistically simulate task outcomes with high accuracy for an agent in a given scenario using PWMs. Considering complex robotic autonomous agents that have to operate and interact in uncertain and stochastic environments, it is necessary for these PWMs to capture the inherent uncertainties of the environment. In this work, we showed how deep generative models, particularly VAEs, can be used to design reliable PWMs that are accurate and well-calibrated in stochastic and partially-observable environments, and able to perform long-horizon forecasting of trajectories. Along with demonstrating the accuracy and reliability of the predicted outcome probabilities from the recurrent VAEs, we showed how forecasted trajectories can be abstracted into outcomes of interest and used to communicate competency.  

There are clear short term extensions of this work, such as comparing the competency statements under varying environmental conditions and policies, both of which can be used to provide additional relevant information to the user. For longer term future work, we plan to develop and study PWMs and competency assessment in more complex and realistic environments. Before getting to this stage, however, we recognize that our developed PWM currently only addresses irreducible aleatoric uncertainty that can be learned from data. Epistemic uncertainty arises from the model weights themselves and is reducible with additional training and data. We will seek to capture and separate the two forms of uncertainties to better calibrate the PWM, particularly in out-of-distribution scenarios. Finally, while the learned PWM in this work does come with a latent space, we have not fully explored it to see what insights it may provide. Thus, we plan to study the latent space to explore how the encoded information can be used to inform our competency assessments about what environmental conditions affect what parts of the agent's dynamics, thus leading to more interpretable autonomous agents.

\addtolength{\textheight}{-3cm} %{-12cm}   % This command serves to balance the column lengths

%%%%%%%%%%%%%%%%%%%%%%%%%%%%%%%%%%%%%%%%%%%%%%%%%%%%%%%%%%%%%%%%%%%%%%%%%%%%%%%%

%%%%%%%%%%%%%%%%%%%%%%%%%%%%%%%%%%%%%%%%%%%%%%%%%%%%%%%%%%%%%%%%%%%%%%%%%%%%%%%%



\section*{ACKNOWLEDGMENT}

This material is based upon work supported by the Defense Advanced Research Projects Agency (DARPA) under Contract No. HR001120C0032. Any opinions, findings and conclusions or recommendations expressed in this material are those of the author(s) and do not necessarily reflect the views of DARPA.
% The preferred spelling of the word ÒacknowledgmentÓ in America is without an ÒeÓ after the ÒgÓ. Avoid the stilted expression, ÒOne of us (R. B. G.) thanks . . .Ó  Instead, try ÒR. B. G. thanksÓ. Put sponsor acknowledgments in the unnumbered footnote on the first page.

%%%%%%%%%%%%%%%%%%%%%%%%%%%%%%%%%%%%%%%%%%%%%%%%%%%%%%%%%%%%%%%%%%%%%%%%%%%%%%%%

% References

% \bibliographystyle{IEEEtran}
\bibliographystyle{plain}
\bibliography{ref}

\end{document}